# Stochastic simulation algorithms for dynamic probabilistic networks

**Keiji Kanazawa, Daphne Koller, Stuart Russell**
Computer Science Division
University of California
Berkeley, CA 94720, USA

## Abstract

Stochastic simulation algorithms such as likelihood weighting often give fast, accurate approximations to posterior probabilities in probabilistic networks, and are the methods of choice for very large networks. Unfortunately, the special characteristics of dynamic probabilistic networks (DPNs), which are used to represent stochastic temporal processes, mean that standard simulation algorithms perform very poorly. In essence, the simulation trials diverge further and further from reality as the process is observed over time. In this paper, we present simulation algorithms that use the evidence observed at each time step to push the set of trials back towards reality. The first algorithm, "evidence reversal" (ER) restructures each time slice of the DPN so that the evidence nodes for the slice become ancestors of the state variables. The second algorithm, called "survival of the fittest" sampling (SOF), "repopulates" the set of trials at each time step using a stochastic reproduction rate weighted by the likelihood of the evidence according to each trial. We compare the performance of each algorithm with likelihood weighting on the original network, and also investigate the benefits of combining the ER and SOF methods. The ER/SOF combination appears to maintain bounded error independent of the number of time steps in the simulation.

## 1 Introduction

Dynamic probabilistic networks or DPNs (Dean and Kanazawa, 1989; Nicholson and Brady, 1992; Kjaerulff, 1992) are a species of belief network designed to model stochastic temporal processes.[1] They do so by using a section of the network called a *time slice* to represent a snapshot of the evolving temporal process. The DPN consists of a sequence of time slices where nodes within time slice $t$ are connected to nodes in time slice $t+1$ as well as to other nodes within slice $t$. Figure 1 shows the coarse structure of a generic DPN. The conditional probability tables (CPTs) for a DPN include a *state evolution model*, which describes the transition probabilities between states, and a *sensor model*, which describes the observations that can result from a given state. Typically, one assumes that the CPTs in each slice do not vary over time. The same parameters therefore will be duplicated in every time slice in the network.

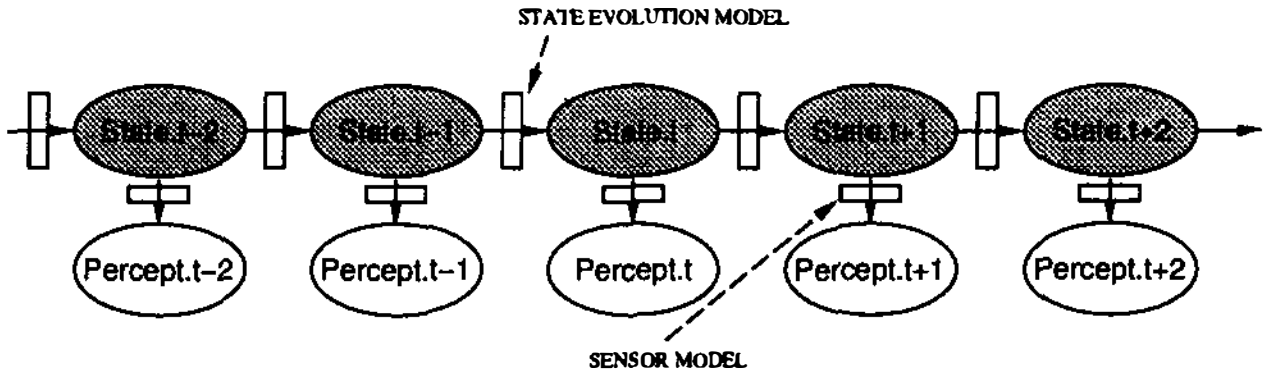

Figure 1: Generic structure of a dynamic probabilistic network. In an actual network, there may be many state and sensor variables in each time slice.

DPNs serve a number of purposes. They can be used for monitoring a partially observable system—for example, Nicholson and Brady used a DPN to track moving robots using light beam sensors. They can be used to project possible future evolutions of the observed system by adding slices into the future. When decision nodes are added, they enable approximately rational decision-making with a limited horizon (Tatman and Shachter, 1990). We have used them for freeway surveillance (Huang *et al.*, 1994) and for controlling an autonomous vehicle (Forbes *et al.*, 1995). In this paper, we concentrate on the use of DPNs for monitoring, i.e., maintaining a probability distribution over the possible current states of the world. Since the correct decision in any partially observable environment depends on this distribution (Astrom, 1965), monitoring is also an essential component of embedded decision-makers.

Exact clustering algorithms for DPNs are described by

[1] Alternative terms include *dynamic belief networks* and *temporal belief networks*.

Kjaerulff (1992). In our applications, we have found that the clustering approach is too expensive and that exact probabilities are not needed. Furthermore, when continuous variables are included, DPNs seldom conform to the structural requirements for CG distributions (Lauritzen and Wermuth, 1989). Hence, exact algorithms are not available. We have therefore investigated the use of stochastic simulation algorithms, which often provide fast approximations to the required probabilities and can be used with arbitrary combinations of discrete and continuous distributions. In the context of DPNs, stochastic simulation methods attempt to approximate the joint distribution for the current state using a collection of "simulated realities," each describing one possible evolution of the environment.

The simplest simulation algorithm is *logic sampling* (Henrion, 1988). Logic sampling stochastically instantiates the network, beginning with the root nodes and using the appropriate conditional distributions to extend the instantiation through the network. Because logic sampling discards trials whenever a variable instantiation conflicts with observed evidence, it is likely to be ineffective in DPN-based monitoring where evidence is observed throughout the temporal sequence.[2]

*Likelihood weighting* (LW) (Fung and Chang, 1989; Shachter and Peot, 1989) attempts to overcome this general problem with logic sampling. Rather than discarding trials that conflict with evidence, each trial is *weighted* by the probability it assigns to the observed evidence. Probabilities on variables of interest can then be calculated by taking a weighted average of the values generated in the population of trials. It can be shown that likelihood weighting produces an unbiased estimate of the required probabilities. The LW algorithm, which we have adapted for the purposes of maintaining beliefs in a DPN as evidence arrives over time, is shown in Figure 2. We use the notation $E_t$ to denote the evidence variables for time slice $t$, and $X_t$ to denote the state variables for time slice $t$. $N$ is the number of samples to be generated, $s_i$ is the $i$th sample, $w_{s_i}$ is its weight, and $T$ is the number of time steps for which the simulation is to be run. *Likelihood*$(E|s)$ denotes the product of the individual conditional probabilities for the evidence in $E$ given the sampled values for their parents in $s$. At each time slice, the current belief for $X_t$ is calculated as the normalized score from the whole sample set.

The use of likelihood weighting in DPNs reveals some problems that require special treatment. The difficulty is that a straightforward application generates simulations that simply ignore the observed evidence and therefore become increasingly irrelevant. Consider a simple example: tracking a moving dot on a 2-D surface. Suppose that the state evolution model is fairly weak—for example, it models the motion as a random walk—but that the sensor is fairly accurate with a very small Gaussian error. Figure 3 illustrates the difficulty. The samples are evolved according to the state evolution model, spreading out randomly over the surface, whereas the object moves along some particular trajectory that is unrelated to the sample distribution. The weighting process will assign extremely low weights to almost all of the samples because they disagree with the sensor observations. The estimated distribution will be dominated by a very small number of samples that are closest to the true state, so the effective number of samples diminishes rapidly over time. This results in large estimation errors. All this occurs despite the fact that *the sensors can track the object with almost no error*! In the case of traffic surveillance, we have discovered that a naive application of likelihood weighting results in a large number of more or less imaginary traffic scenes that bear almost no relation to what is actually happening on the road.

**procedure** LIKELIHOOD-WEIGHTING()

  **loop for** $i = 1 \ldots N$
    $w_{s_i} \leftarrow 1.0$
  **loop for** $t = 0 \ldots T$
    Instantiate $E_t$
    **loop for** $i = 1 \ldots N$
      Add sample of $X_t$ to $s_i$
      $w_{s_i} \leftarrow w_{s_i} \times$ *Likelihood*$(E_t \mid s_i)$
      Add $w_{s_i}$ to score for sampled values of $X_t$

Figure 2: The Likelihood Weighting algorithm.

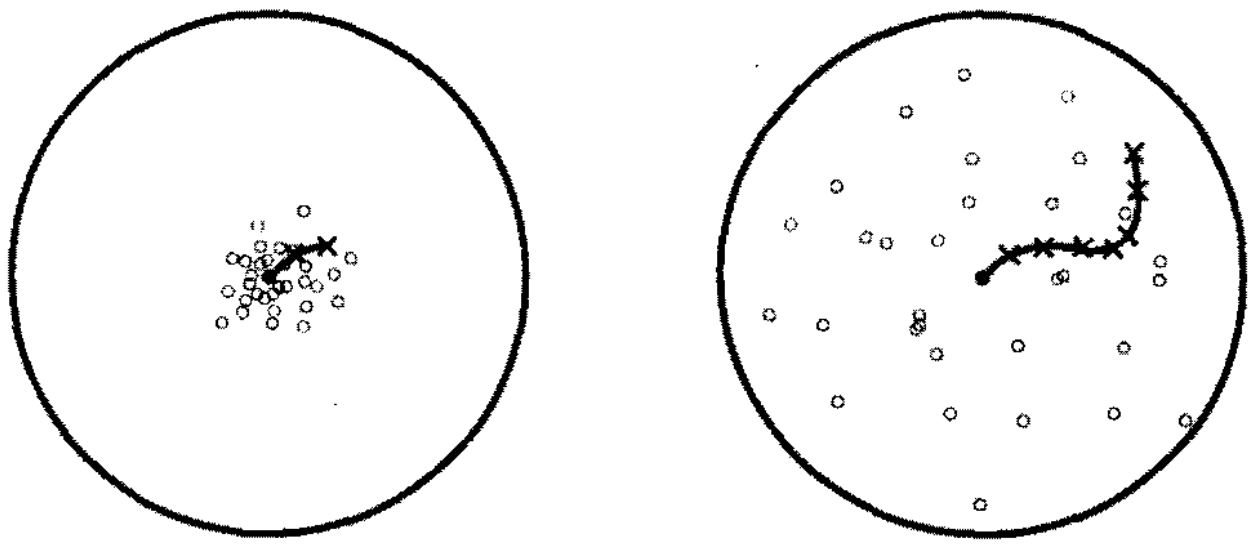

Figure 3: A simple 2-D monitoring problem. An object starts in the centre of the disc and follows the path shown by the solid line. Sensor observations are shown by crosses. The small circles show a snapshot of the population of samples generated by a naive application of likelihood weighting. Snapshots for $t = 2$ and $t = 7$ are shown.

Clearly, we need algorithms that use the current sensor values to reposition the sample population closer to reality rather than allowing them to evolve as if no sensor values were available. Section 2 describes a simple method (*evidence reversal*) for restructuring the DPN so that likelihood weighting has the desired effect. Section 3 describes a related method (*survival of the fittest*) that uses the likelihood weights to prefer-entially propagate the most likely samples, and shows how this can be combined with evidence reversal. Section 4 describes an experimental comparison of these techniques with naive LW.

[2]On the other hand, logic sampling is extremely effective for projection, because no evidence is observed in future slices.

## 2 Evidence reversal

It has long been known that stochastic simulation algorithms are quite effective if the network contains no evidence (Dagum and Luby, 1993). The same argument can be used to show that if all the evidence in a network is at the root nodes, approximating the probabilities in the rest of the network is computationally tractable. This explains the appeal of logic sampling for projection, but is not directly applicable to the monitoring problem (where evidence is obtained for every time slice). We can force the evidence to be at the root nodes of any network simply by reversing all the arcs using Shachter's (1986) transformations, but doing this to an $n$-slice DPN results in an exponential blowup. As a compromise, we can do some judiciously selected arc reversals as suggested by Fung and Chang. In the specific case of DPNs, we can take advantage of the fact that each sample, once it instantiates variables in time slice $t-1$, d-separates all preceding time slices from the state at time $t$. We then simply reverse the arcs within slice $t$, so that the evidence at $t$ and the state at $t-1$ become the parents of the state at time $t$. This is shown in schematic form in Figure 4.

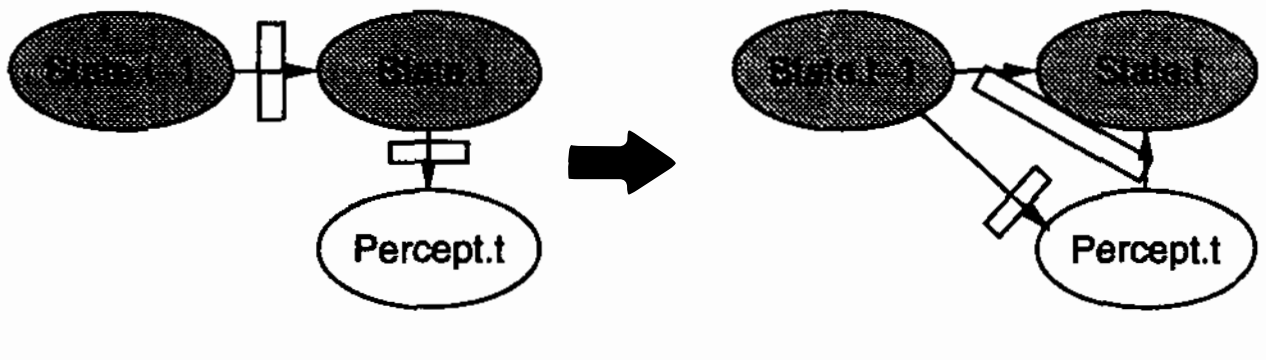

Figure 4: Schematic diagram of the evidence reversal transformation for DBNs.

The process is then as follows. For each time slice, we have some number $k$ of fully specified states along with their weights.

1. Reverse the arcs from evidence to state at time $t$; the state variables at time $t-1$ are now parents of the evidence at time $t$.
2. Use the evidence at time $t$ to adjust the weights of the samples at time $t-1$, as in standard likelihood weighting.
3. Propagate each sample at time $t-1$ through the modified state-evolution model which uses the evidence at time $t$ (as obtained in the arc-reversed time slice).

In ER, the current evidence is a parent of the current state; therefore, it can influence the process of extending the samples to the state variables at $t$. In particular, in the 2-D tracking example shown in Figure 3, all the samples will stay closely clustered around the observed position of the object because the accurate sensor readings will dominate the weak state evolution model in the conditional distribution for generating the new samples.

## 3 Survival of the fittest

The problem with the naive application of sampling algorithms can also be viewed as one of resource allocation. The samples are a constrained resource, and should be allocated in the state space to try to "fit" the actual joint distribution as well as possible. Samples that have wandered off into totally imaginary scenarios should not be propagated, since they do not contribute enough to the estimation of the desired probabilities. The idea of survival-of-the-fittest (SOF) sampling is to preferentially propagate forward in time those samples that have high likelihood for the observed evidence. The SOF process keeps a fixed number of samples, but generates the sample population for time slice $t$ by a weighted random selection from the samples at time $t-1$, where the weight is given by the likelihood for the evidence observed at time $t$. This idea is closely related to the use of fitness-related propagation in genetic algorithms and the sample-repositioning method used in randomized "go with the winners" algorithms (Aldous and Vazirani, 1994).

The SOF approach can also be understood as a slice-by-slice likelihood weighting process. Rather than using the samples to provide an approximation to the joint probability distribution over the entire (multi-slice) network, we only use them to propagate the belief state—the joint probability distribution over the state—from one time slice to the next. More precisely, the weighted samples at time $t-1$ are an approximation to the belief state at time $t-1$. We can then use *that approximate belief state* as our starting point for the sampling at the next time slice. That is, we sample each state according to its weight, as defined by our current (likelihood weighted) samples. These samples are in turn weighted using the evidence at time $t$, and provide an approximation of the belief state at time $t$. Note that the probability of sampling a given state at time $t$ is given just by the likelihood for the evidence at time $t$, and not by the accumulated likelihood for all evidence up to and including time $t$ (as would be the case in standard likelihood weighting). This is because the sample population at time $t-1$ in SOF *already* reflects the evidence up to time $t-1$ through the process of preferential propagation. The algorithm is shown in Figure 5.

SOF clearly provides some improvement over likelihood weighting in general, but does not take advantage of the sensor values in quite the same way as ER. In the context of the 2-D tracking problem, SOF will multiply the samples closest to the actual track so that almost the entire population consists of "reasonable" samples and will never spread out over the entire surface. However, the samples will spread out

```
procedure SOF()

  loop for t = 0 ... T
    Instantiate E_t
    loop for i = 1 ... N
      Add sample of X_t to s_i
      w_{s_i} ← Likelihood(E_t | s_i)
      Add w_{s_i} to score for sampled values of X_t
    Repopulate sample set by randomized
      selection weighted by w_{s_i}
```

Figure 5: The Survival-of-the-Fittest algorithm.

by an amount related to the uncertainty in the state evolution model, regardless of how accurate the sensor model is. Fortunately, the advantages provided by ER and SOF can be combined into an ER/SOF hybrid, simply by applying SOF to the ER sampling process. That is, rather than propagating all the slice-$t-1$ samples at step 3 in the ER algorithm, we use the SOF technique to focus on the ones that are most likely. That is, we sample from the distribution obtained in step 2 of the ER algorithm, and then propagate those through the modified state-evolution model.

## 4 Empirical results

In this section, we report on some simple experiments we carried out to confirm the intuitive ideas presented above. The network used in our experiments has the same topology as the network shown in Figure 1.[3] The aim is to investigate the problem of sample population divergence over time, and to show that ER and SOF mitigate the problem. We measure the average absolute error in the marginal probabilities of the state variables of a time slice as a function of $t$—that is, the $x$-axis measures time in the simulated environment.

Figure 6 shows the error behaviour for LW over 50 time steps for 25, 100, 1000, and 10000 samples, averaged over 50 randomly generated sets of evidence. The results clearly show that LW fails dramatically even on this very simple network. The problem is that as any given sample is propagated over time, sooner or later it will sample a state value that makes the observed evidence impossible (for each state value in our network, one of the four observation values in not possible). After sufficiently many steps, *all* the samples end up with weight 0, at which point we assign an error of 1.0. Thus, after 39 steps with 25 samples, all the samples are extinguished in all 50 cases. Multiplying the number of samples only delays the inevitable by a small number of steps.

Figure 7 shows the corresponding error behaviour for ER. Note that the scale of the $y$-axis is increased by 10. Thus, the error remains well within the acceptable range. It does, however, show a slow increase over time. It is possible that the error asymptotes as $t \to \infty$, but we have not yet run those experiments.

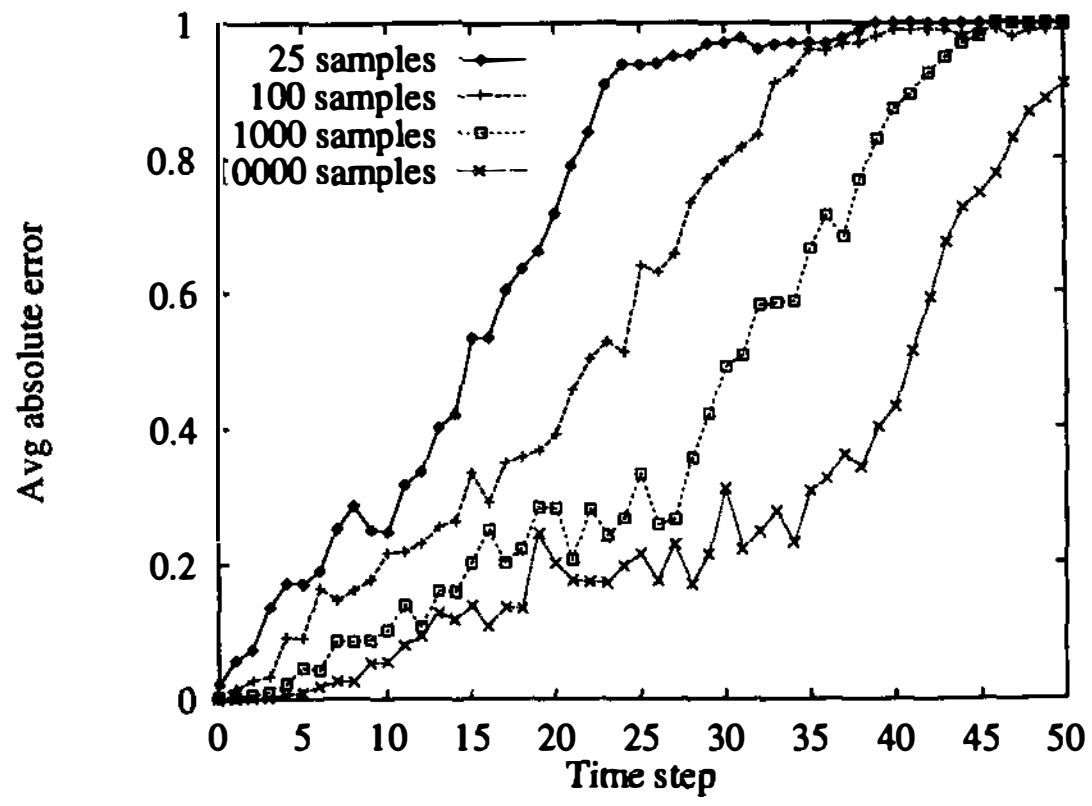

Figure 6: Performance of LW: Graph showing the average absolute error in the marginal probabilities of the state variables of a time slice as a function of $t$, averaged over 50 randomly generated evidence cases.

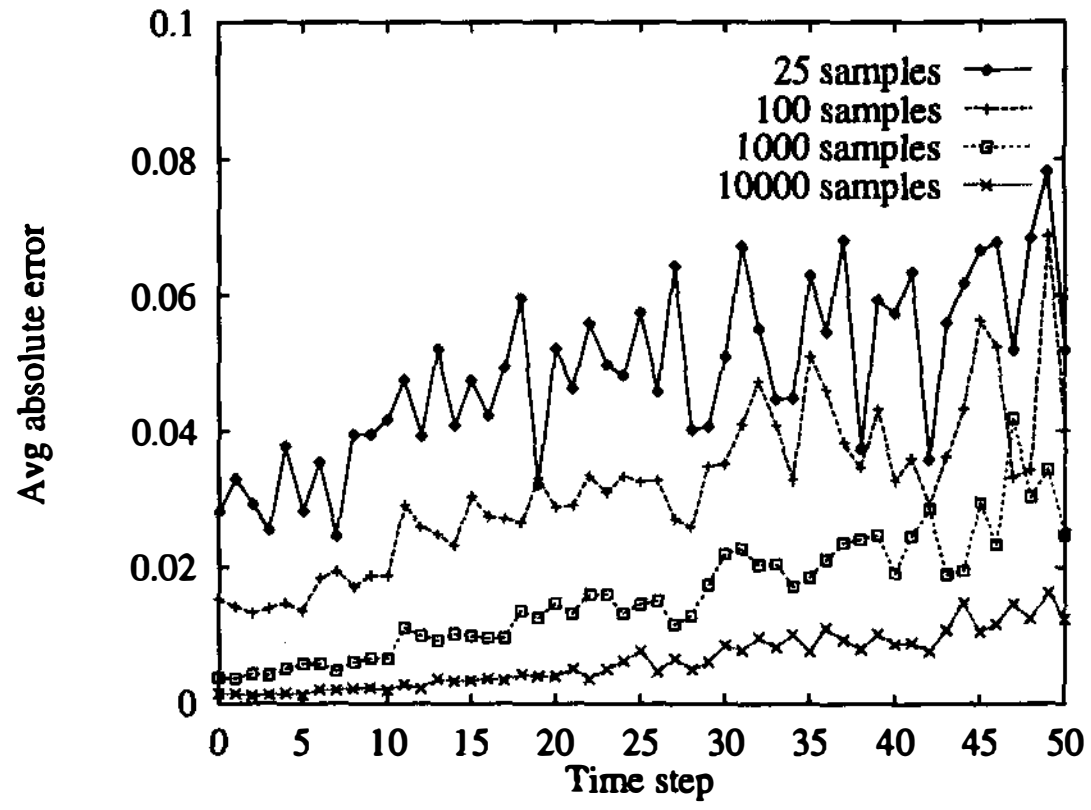

Figure 7: Performance of ER: Graph showing the average absolute error in the marginal probabilities of the state variables of a time slice as a function of $t$, averaged over 50 randomly generated evidence cases.

Figures 8, 9, and 10 show the performance of SOF and ER/SOF, compared with ER, for 25, 100, and 1000 samples respectively. The results show that SOF is an effective mechanism for maintaining bounded error over time. Although SOF on its own shows somewhat higher error than ER, as one would expect, the combination of ER and SOF shows low error for all time steps and shows no sign of diverging at all.

Finally, Figure 11 shows the performance of ER, SOF, and ER/SOF as a function of the number of samples for the range 50 to 1000 samples. The graph gives the average absolute error in the marginal probabilities of the state variable at $t = 50$. The graphs show that SOF seems to benefit much less from additional samples than ER—in fact, the curve is almost flat. Currently, our theoretical analysis of the algorithm is not sufficiently advanced to explain this phenomenon.

[3]We are currently working to generate similar experimental data for our traffic surveillance networks.

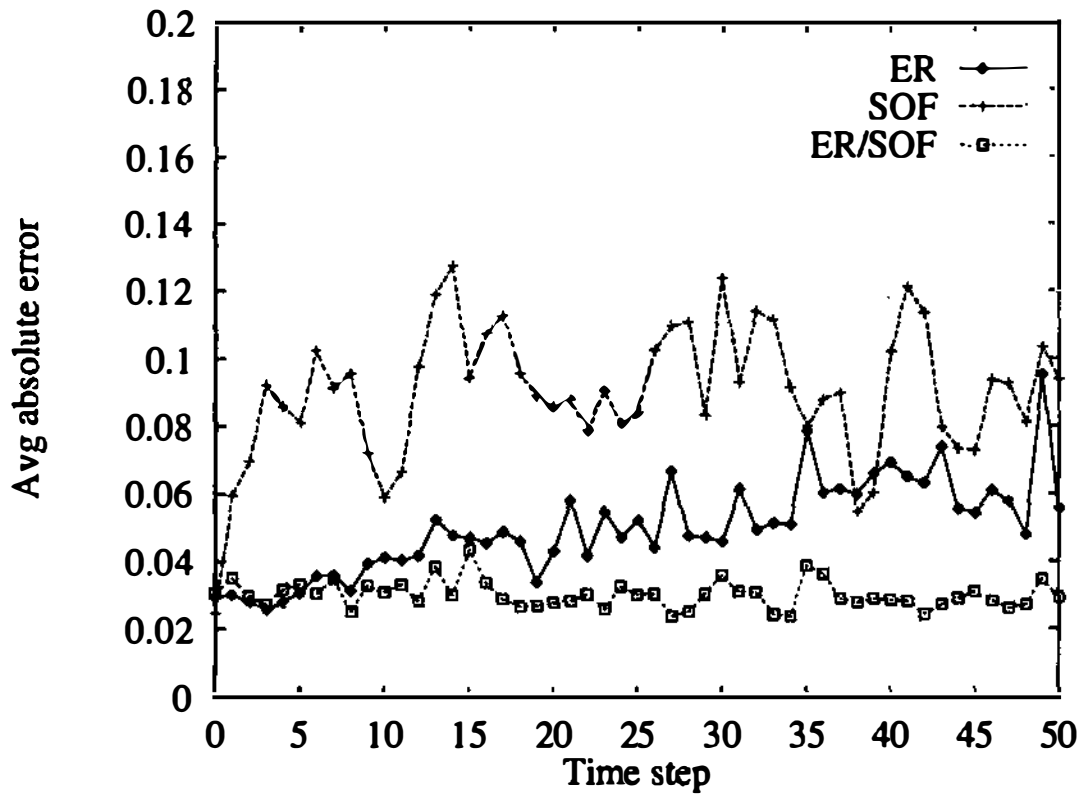

Figure 8: Performance of ER, SOF, and ER/SOF: Graph showing the average absolute error in the marginal probabilities of the state variables of a time slice as a function of $t$, averaged over 50 randomly generated evidence cases, for 25 samples.

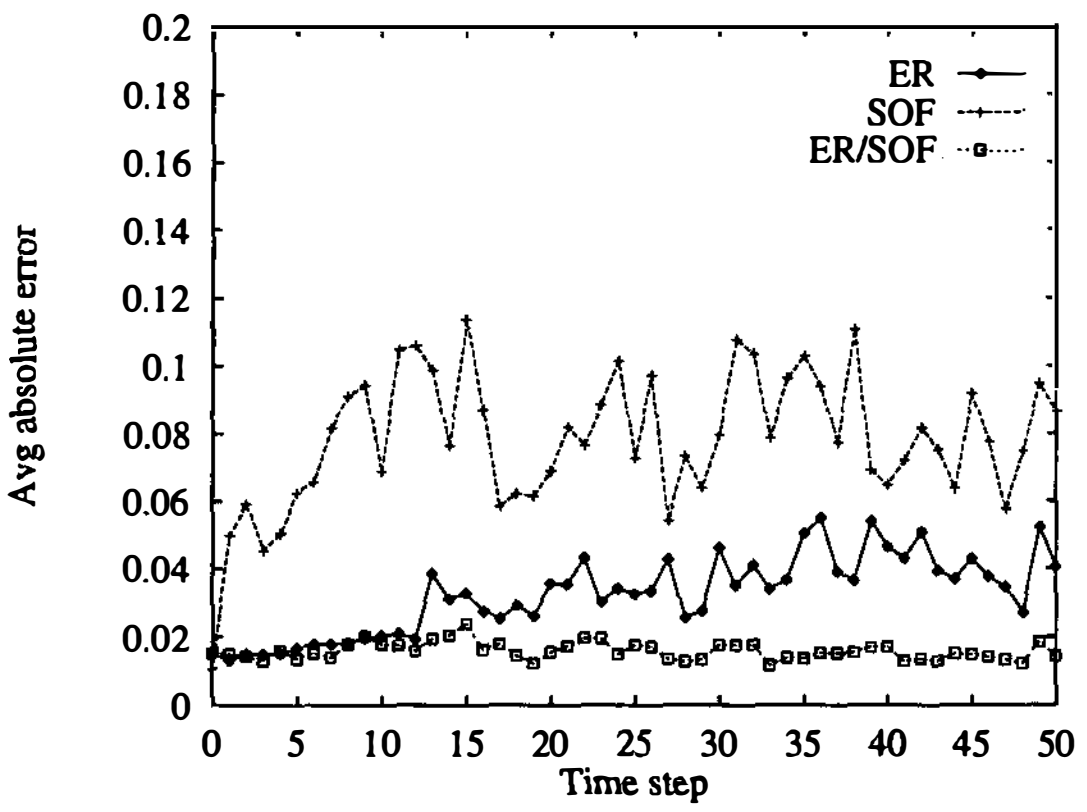

Figure 9: Performance of ER, SOF, and ER/SOF: Graph showing the average absolute error in the marginal probabilities of the state variables of a time slice as a function of $t$, averaged over 50 randomly generated evidence cases, for 100 samples.

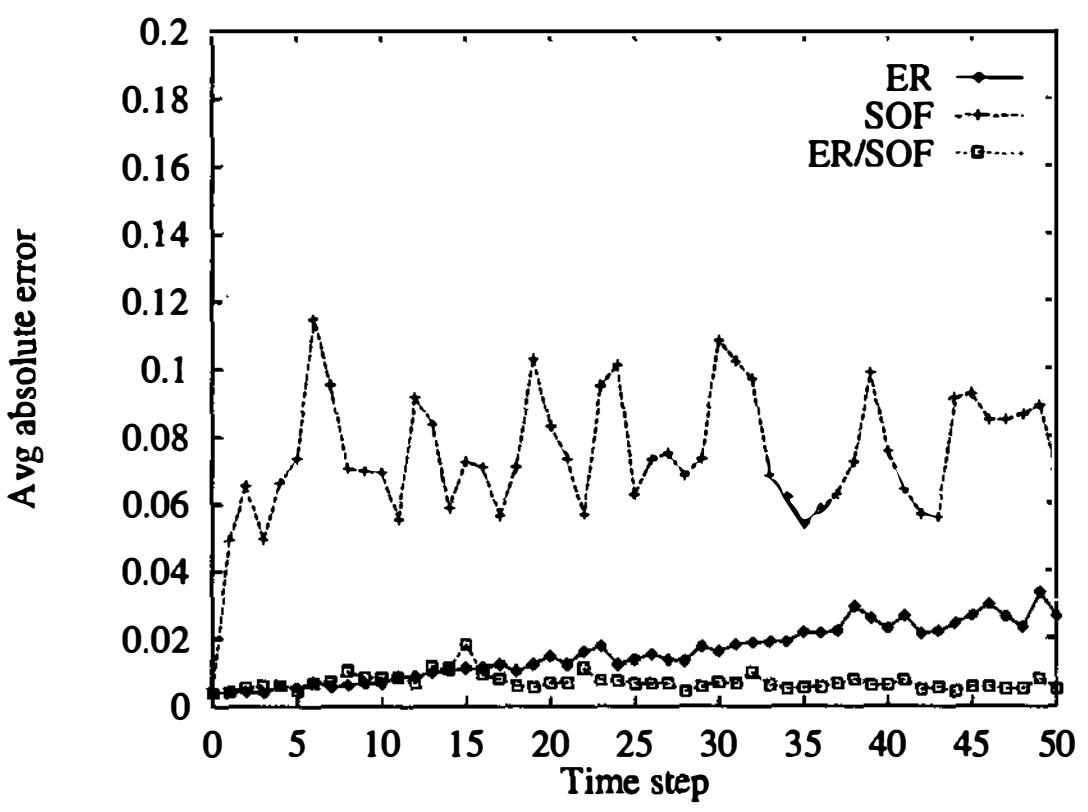

Figure 10: Performance of ER, SOF, and ER/SOF: Graph showing the average absolute error in the marginal probabilities of the state variables of a time slice as a function of $t$, averaged over 50 randomly generated evidence cases, for 1000 samples.

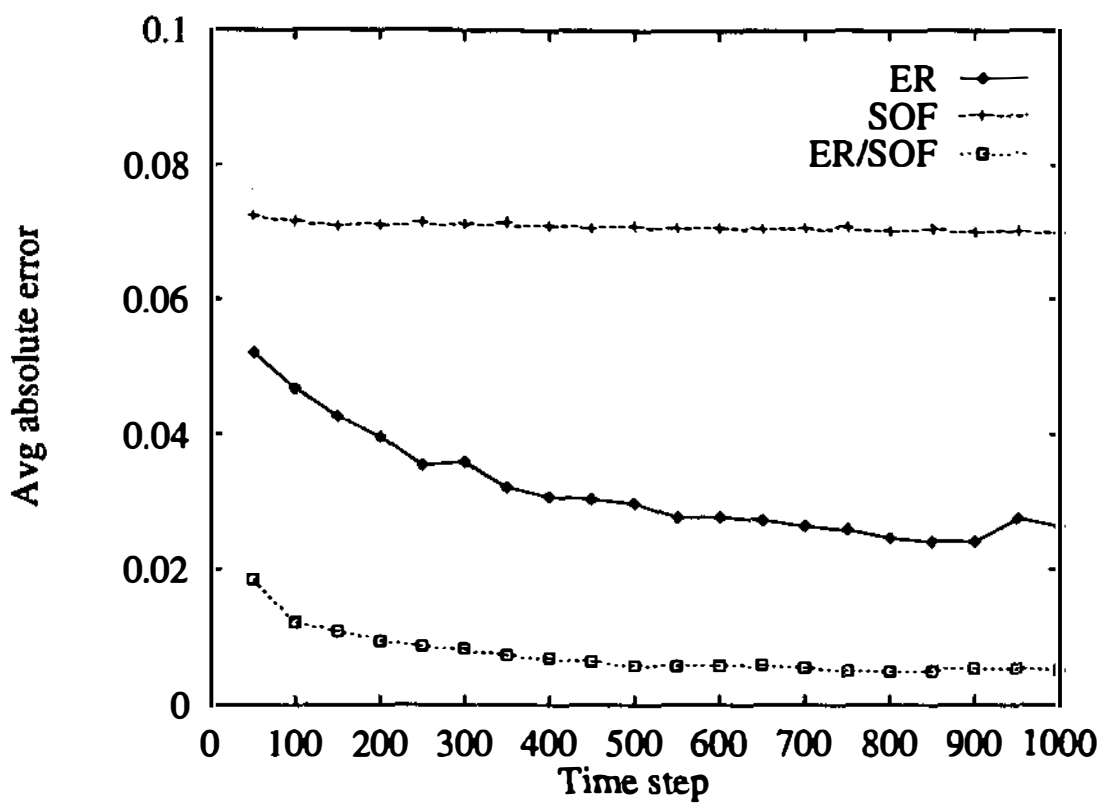

Figure 11: Performance of ER, SOF, and ER/SOF as a function of the number fo samples: Graph showing the average absolute error in the marginal probabilities of the state variables for time slice $t = 50$, averaged over 50 randomly generated evidence cases.

## 5 Conclusion and further work

We have presented two very simple and intuitive improvements that make the likelihood weighting technique effective for dynamic probabilistic networks. Early experimental results confirm our intuitions. In particular, the error for SOF and ER/SOF seems to be independent of the number of time steps in the simulation. This is an absolute requirement for monitoring applications such as traffic surveillance, where inference continues over many days of real time.

Further work needs to be done to establish the theoretical properties of the algorithms. The most obvious issue is whether these approaches are unbiased: do they converge to the right answer as the number of samples grows to infinity. ER is clearly unbiased, because it just an application of likelihood weighting to a modified network structure. It seems fairly straightforward to show that SOF (and therefore ER/SOF) converge to the correct values in the large-sample limit using standard probabilistic techniques.

We would also like to investigate the expected error as a function of sample size for LW, ER, SOF, and ER/SOF. This should be fairly simple for specific network structures such as that shown in Figure 1. Understanding the algorithms' behaviour for general DPNs is more difficult. Intuitively, the improvement of ER and SOF is more pronounced in those cases where the evidence gives us a lot of information about the state. At one extreme, if the sensor model is completely accurate, ER will be completely accurate with only a single sample. The behavior of SOF in these circumstances will also depend on the behavior of the state-evolution model. If this is fairly well-behaved, it appears that SOF will also do well. At the other extreme, if the sensor model is just noise, neither approach seems to provide an advantage over LW. We hope to analyze the improvement of these algorithms using such quantities as (1) the distance (in terms of relative entropy) between the belief-state distribution at time $t$ and at time $t+1$, and (2) the amount of information (in terms of entropy) obtained by considering the sensors.

Finally, SOF is a technique that can be applied to arbitrary networks, not just DPNs. It would be interesting to see if it provides consistently better results than LW for general networks. Since LW is currently the best algorithm known for very large networks, this would be a useful development.